# COMPARISON STUDY FOR CLONAL SELECTION ALGORITHM AND GENETIC ALGORITHM

Ezgi Deniz Ülker[1] and Sadık Ülker[2]

[1]Department of Computer Engineering, Girne American University, Karmi Campus, Girne, Mersin 10 TURKEY
`ezgideniz@gau.edu.tr`
[2]Department of Electrical and Electronics Engineering, Girne American University, Karmi Campus, Girne, Mersin 10 TURKEY
`sulker@gau.edu.tr`

## ABSTRACT

*Two metaheuristic algorithms namely Artificial Immune Systems (AIS) and Genetic Algorithms are classified as computational systems inspired by theoretical immunology and genetics mechanisms. In this work we examine the comparative performances of two algorithms. A special selection algorithm, Clonal Selection Algorithm (CLONALG), which is a subset of Artificial Immune Systems, and Genetic Algorithms are tested with certain benchmark functions. It is shown that depending on type of a function Clonal Selection Algorithm and Genetic Algorithm have better performance over each other.*

## 1. INTRODUCTION

Many heuristic optimization algorithms have been developed and solved in the solution of certain benchmark functions as well as being applied to common optimization problems over the years. Most of these algorithms are inspired by biological studies, and some of them developed through the study of immunology and genetics.

Immune Algorithm, which is developed from immunology, is studied with various settings and got some modifications over the years as well as being applied to various common day problems [1]-[5]. Clonal Selection Algorithm is a special class of Artificial Immune System which uses the clonal selection part of the Artificial Immune Systems as a main mechanism. This algorithm was initially proposed to solve nonlinear functions by De Castro and Van Zuben in 2000 [7]-[8].
The comparative studies have always been of interest to find which algorithm is more superior to the other. Comparative study of Artificial Bee Colony algorithm was done by Karabo a and Akay [9], comparative study of memetic algorithm is done by Merz and Freisleben[10]. In this work the performance of Clonal Selection Algorithm and Genetic Algorithm. Similar work was done before [11] however our aim is to extend this work to different types of functions with different parameters and find a clear performance relationship between the type of a function and the algorithm.
.

            107

## 2. ALGORITHMS

### 2.1. Clonal Selection Algorithm

Immune Algorithm is derived through the study of immune response. In short, it models how antibodies of the immune system learn adaptively the features of the intruding antigen and act upon. Clonal Selection Algorithm is a special class of Artificial Immune Systems. In this work, CLONALG algorithm which was originally proposed by De Castro and Van Zuben [7] is used.

The algorithm starts by defining a purpose function f(x) which needs to be optimized. Some possible candidate solutions are created, antibodies will be used in the purpose function to calculate their affinity and this will determine the ones which will be cloned for the next step. The cloned values are changed, mutated with a predefined ratio and the affinities are recalculated and sorted. After certain evaluations of affinity, affinity with the smallest value is the solution closest to our problem.

Clonal Selection Algorithm can be listed as follows:

*Step 1*: Generate a set of antibodies (generally created in a random manner) which are the current candidate solutions of a problem.
*Step 2*: Calculate the affinity values of each candidate solutions.
*Step 3*: Sort the antibodies starting from the lowest affinity. Lowest affinity means that a better matching between antibody and antigen.
*Step 4*: Clone the better matching antibodies more with some predefined ratio.
*Step 5*: Mutate the antibodies with some predefined ratio. This ratio is obtained in a way that better matching clones mutated less and weakly matching clones mutated much more in order to reach the optimal solution.
*Step 6*: Calculate the new affinity values of each antibody.
*Step 7*: Repeat Steps 3 through 6 while the minimum error criterion is not met.

### 2.2. Genetic Algorithm

Biologists have worked with genetic systems in order to find the information that passed from parents to children through the genes. Mendel first worked with genetics and his works published first in 1865 and then in English in 1901 [12]. Holland can be considered as the father of Genetic Algorithms since his paper in 1975 [13].

Genetic Algorithm is developed [14], and many researchers from various fields adapted Genetic Algorithms into their optimization problems in engineering, biology, economics, physics and many other fields. Genetic Algorithms have also been used in solving multi-objective optimization problems since Shaffer proposed in 1985 [15]. Genetic Algorithms has been used in applications for a long time. An example work was done by Nongmeikapam, and Bandyopadhyay [16].

The steps of Genetic Algorithm can be listed as follows:

*Step 1*: Generate a set of chromosomes (created in a random manner), and these are the current candidate solutions for the given problem.
*Step 2*: Calculate the fitness values of each chromosome.
*Step 3*: Sort the chromosomes starting from the lowest fitness (i.e. better solution).
*Step 4*: Clone the better chromosomes in order to spread good attributes through the population.

*Step 5*: Perform a crossover operation for the parent chromosomes to obtain new offspring population.
*Step 6:* Mutate the chromosomes with a small predefined ratio.
*Step 7*: Calculate the new fitness values of each chromosome.
*Step 8*: Repeat Steps 3 through 7 while the minimum error criterion is not met.

## 3. PROBLEM DEFINITON AND FUNCTIONS

In this work, our aim was to apply Clonal Selection Algorithm and Genetic Algorithm to certain benchmark functions and examine their performances with each other. The performances are evaluated by comparing convergence to the optimum value and number of iterations required to obtain a solution to our problem.

The benchmark functions that we worked on are as follows: Sphere Function, Rastrigin's Function, Ackley's Function, Modified Sinusoidal Function, Sum of Different Power Function, and Non-generalized Schwefel's Function. The graphs for 2 dimensional version of each of these functions are shown in Fig.1 through Fig.6.

### 3.1. Sphere Function

Sphere function is a quadratic function of n dimensions. This function is continuous, convex and unimodal. For these reasons almost all the optimization algorithms can find the global optimum solution. The general function for n dimensions can be given in equation (1):

$$f(x) = \sum_{i=1}^{n} x_i^2 \qquad (1)$$

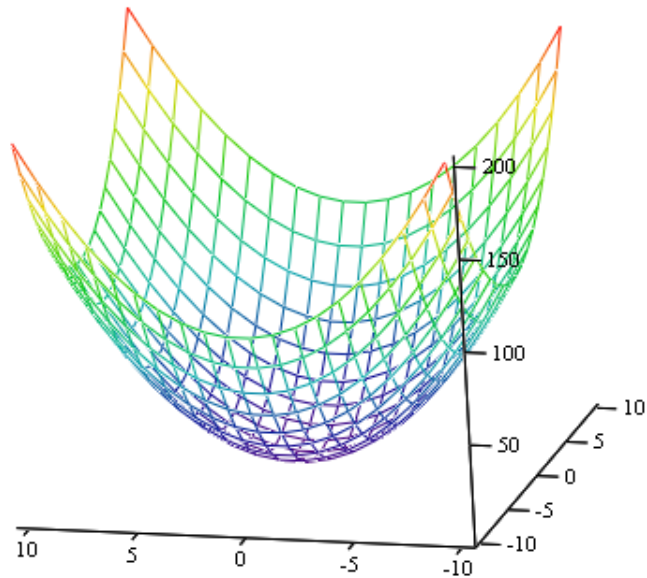

Figure 1. Sphere Function in two dimensions

This function is evaluated by generating numbers in the range –100 to 100 and for 10 dimensions. The global minimum value for this function is 0 and it occurs when all the variables are 0.

## 3.2 Rastrigin's Function

The Rastrigin's function is based on the sphere model with the addition of cosine modulation in order to generate many local minimums. The Rastrigin's function is highly multimodal. The function for n dimensions can be given in equation (2):

$$f(x) = 10n + \sum_{i=1}^{n} \left[ x_i^2 - 10\cos(2\pi x_i) \right] \qquad (2)$$

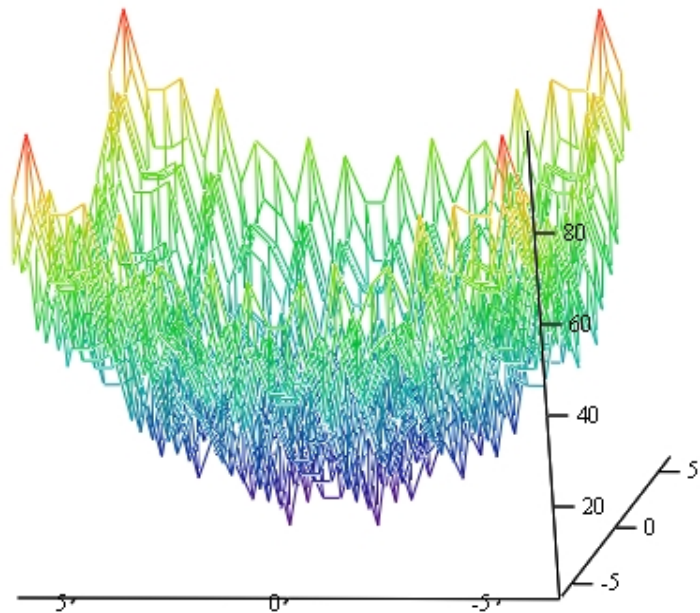

Figure 2: Rastrigin's Function in two dimensions

This function is evaluated by generating numbers in the range –5.12 to 5.12 and with 10 dimensions. The global minimum value is 0 and it occurs when all the variables are 0.

## 3.3 Ackley's Function

The Ackley problem is a benchmark function which is multimodal, and for n dimensions it is given by equation (3):

$$f(x) = -20\exp\left(-0.2\sqrt{\frac{1}{n} \cdot \sum_{i=1}^{n} x_i^2}\right) - \exp\left(\frac{1}{n}\sum_{i=1}^{n}\cos(2\pi x_i)\right) + 20 + e$$

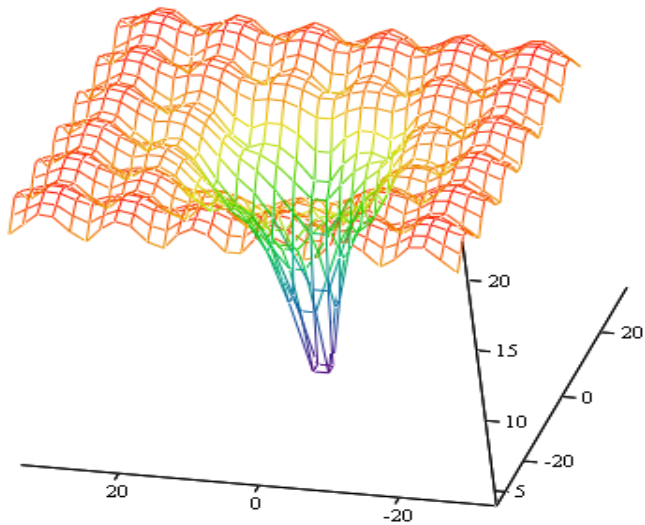

Figure 3. Ackley's function in two dimensions

This function is evaluated by generating numbers in the range –32 to 32 and with 10 dimensions. The global minimum value is 0 and it occurs when all the variables are 0.

### 3.4. Modified Sinusoidal function

Non-modified sinusoidal function has many global minimum values which correspond to many different points. To refrain from these points, sinusoidal function is modified and reduced to one global minimum point. This function is evaluated by generating numbers in the range 0 to 6 and with 10 dimensions. The function in n dimensions is given by equation (4):

$$f(x) = \sum_{i=1}^{n} \sin(x_i) + n \qquad (4)$$

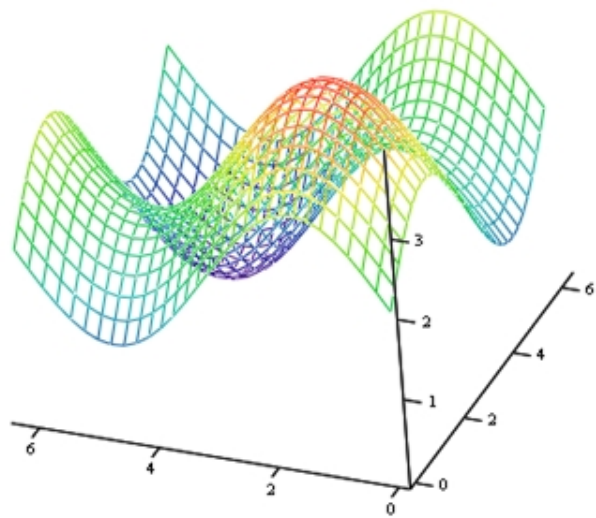

Figure 4. Modified Sinusoidal function in two dimensions

The global minimum value for ten variables is 0 and it occurs when all the variables are 4.714.

### 3.5. Sum of Different Powers function

This function is a smooth shaped unimodal benchmark function. The equation for n dimensions is given by equation (5):

$$f(x) = \sum_{i=1}^{n} |x_i|^{i+1} \tag{5}$$

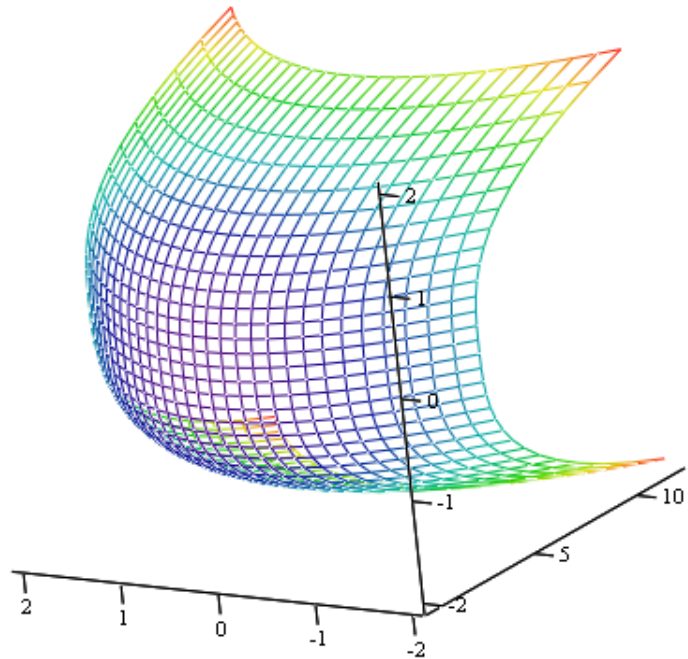

Figure 5. Modified Sinusoidal function in two dimensions

This function is evaluated by generating numbers in the range –2.048 to 2.048 with 10 dimensions. The global minimum value is 0 and it occurs when all the variables are 0.

### 3.6. Non-Generalized Schwefel's function

Non-generalized Schwefel's function has one global peak and many local peak points, for n dimensions it is given by the following equation (6):

$$f(x) = \sum_{i=1}^{n} |x_i| + \prod_{i=1}^{n} |x_i| \tag{6}$$

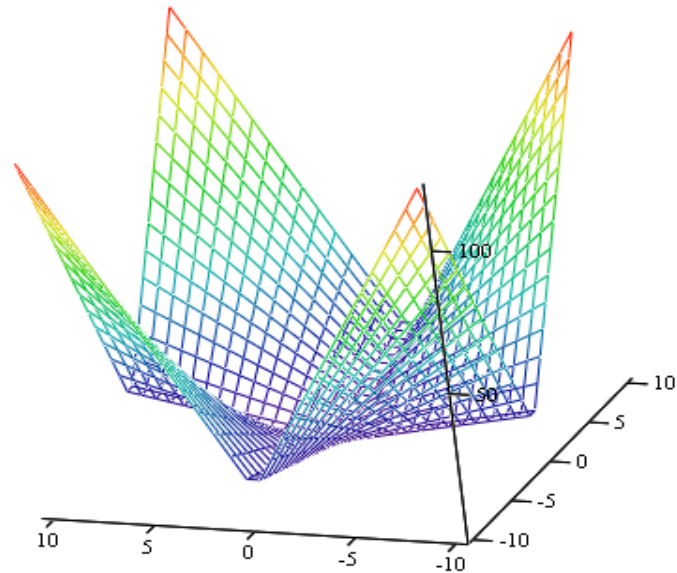

Figure 6. Non-Generalized Schwefel's function in two dimensions

The function is evaluated by generating numbers in the range –10.0 and 10.0 with 10 dimensions. The global minimum value is 0 and it occurs when all the variables are 0.

## 4. RESULTS

Functions explained above are the benchmark test functions used for evaluating the Clonal Selection Algorithm and Genetic Algorithm. While the tests are performed, search space is kept same for each of the problems. The search space consists of 40 rows of being either 0 or 1 (i.e. as bits) and 200 columns. Each row holds 10 variables with 20 bits each. During the testing the benchmark functions with Clonal Selection and Genetic Algorithm parameters are selected as cloning rate and mutation rate for both algorithms and they are changed for each run. Performances of algorithms are shown in between Fig.7 and Fig.12 for each function.

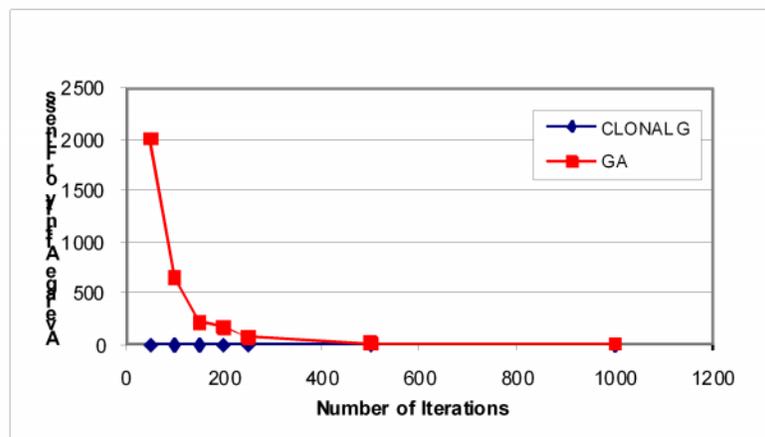

Figure 7. Responses for Sphere Function

Mutation rates are different for two algorithms because mutation is the main driving parameter in Clonal Selection Algorithm however crossover and mutation together are the main components in Genetic Algorithm. Cloning rate is very important for both algorithms. Genetic Algorithm normally uses Roulette Wheel Selection in natural selection part however in this work instead of Roulette Wheel Selection various cloning rates were used to give higher probability to the better fitness values in Genetic Algorithm.

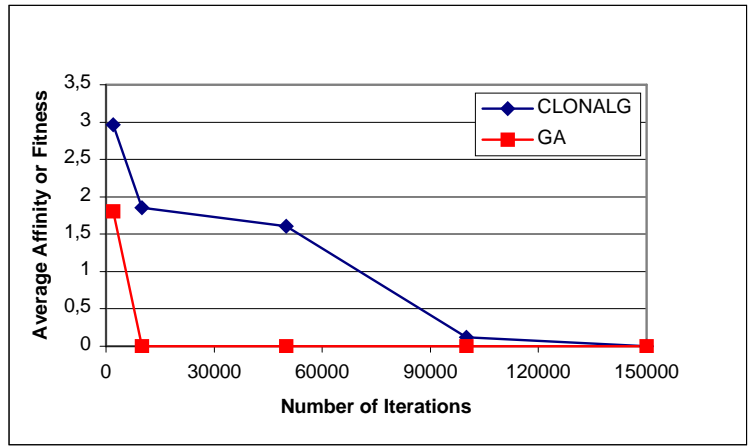

Figure 8. Responses for Rastrigin's Function

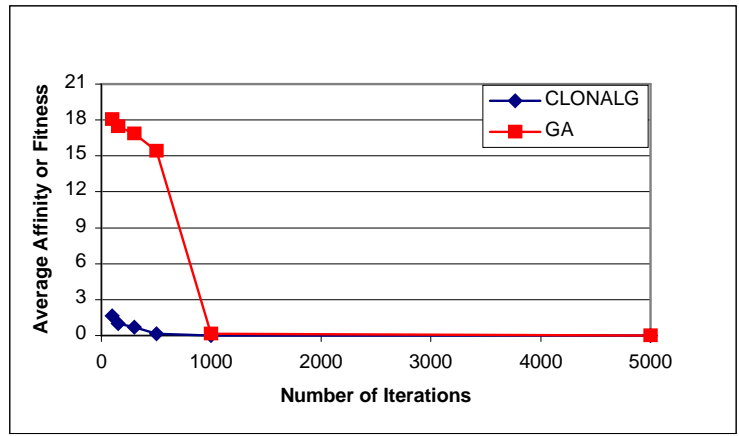

Figure 9. Responses for Ackley's Function

Many different rates of parameters are used in both of the algorithms but some rates worked better than the others. Rates that gave better results are selected for comparison. Three different cloning and mutation rates are used in Clonal Selection Algorithm and Genetic Algorithm.

In the Clonal Selection Algorithm, four best affinity values are selected and cloned with a predefined ratio. Then antibodies are mutated with the given rates. The cloning rate is determined in a way that better matching antibodies are cloned much more and weakly matching antibodies are cloned less. With a similar logic better matching antibodies are mutated less and weakly matching antibodies are mutated more. Similar approach is used in Genetic Algorithm.

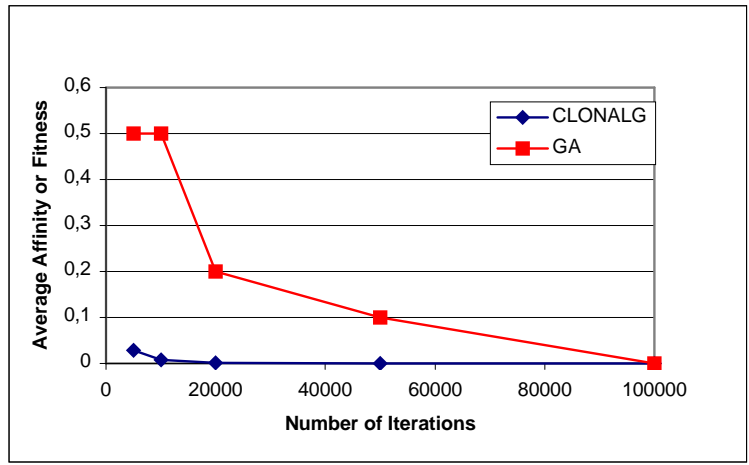

Figure 10: Responses for Modified Sinusoidal Function.

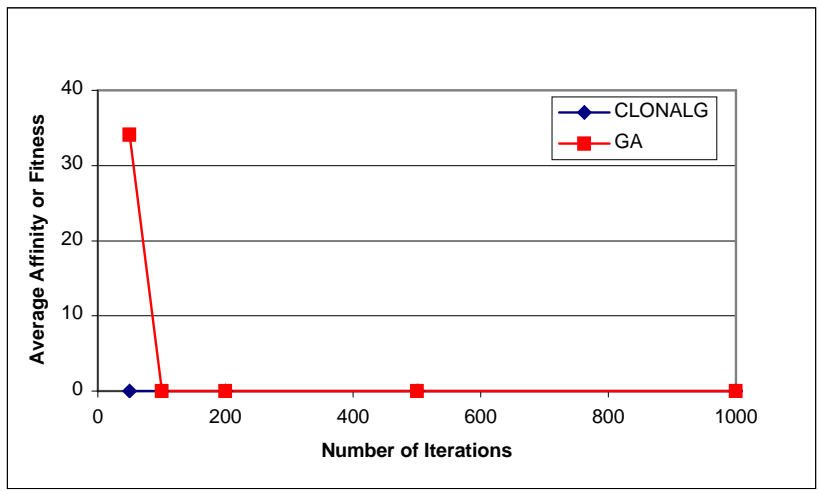

Figure 11: Responses for Sum of Different Power Function

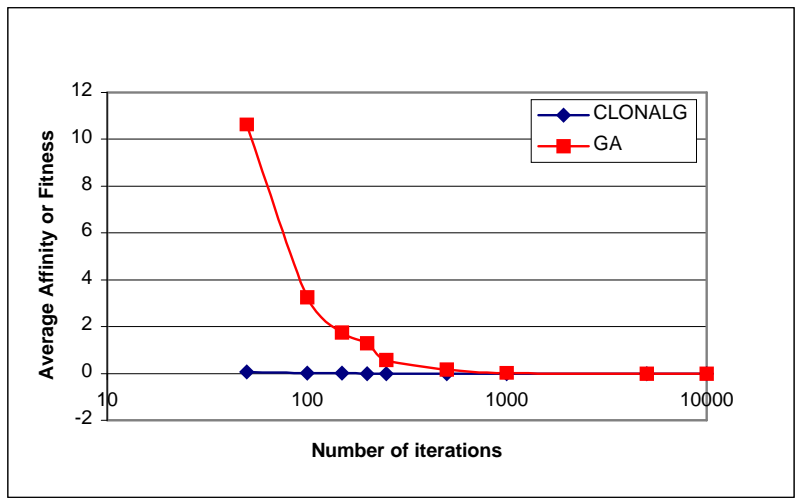

Figure 12: Responses for Non-Generalized Schwefel's Function

Three different sets for cloning were done in Table 1. The table shows the clone rates for Clonal Selection and Genetic Algorithm.

TABLE 1. DEFINITION OF SETS FOR THE CLONE RATES

|  | Number of cloned antibody/chromosome in each set | | |
| --- | --- | --- | --- |
|  | Set 1 | Set 2 | Set 3 |
| **First best Antibody/Chromosome** | 15 | 17 | 18 |
| **Second best Antibody/Chromosome** | 7 | 8 | 9 |
| **Third best Antibody/Chromosome** | 5 | 5 | 6 |
| **Fourth best Antibody/Chromosome** | 3 | 4 | 4 |

The mutation rates that are applied for each set are composed of three groups; Group 1, consisting

TABLE II. COMPARISON OF ALGORITHMS WITH THE BEST PARAMETERS

| Function | Type | Clonal Selection Algorithm | | | | Genetic Algorithm | | | |
| --- | --- | --- | --- | --- | --- | --- | --- | --- | --- |
|  |  | Sets for clone rate | Group number for mut. rate | Proximity | Avg. Number of Iterations | Sets for clone rate | Mut. Rate | Proximity | Avg. Number of Iterations |
| Sphere | Unimodal | 2 | 1 | $6.95 \times 10^{-7}$ | 399 | 2 | 0.005 | $4.84 \times 10^{-7}$ | 3589 |
| Rastrigin's | Highly Multimodal | 3 | 3 | $8.51 \times 10^{-5}$ | 135226 | 1 | 0.001 | $4.73 \times 10^{-5}$ | 4945 |
| Ackley's | Multimodal | 2 | 1 | $8.59 \times 10^{-4}$ | 417 | 1 | 0.001 | $2.37 \times 10^{-4}$ | 2310 |
| Modified Sinusoidal | Highly Multimodal | 3 | 2 | $9.71 \times 10^{-4}$ | 14488 | 1 | 0.001 | $8.13 \times 10^{-4}$ | 10474 |
| Sum of Different Powers | Unimodal | 3 | 1 | $6.21 \times 10^{-6}$ | 53 | 1 | 0.005 | $4.59 \times 10^{-6}$ | 375 |
| Non-generalized Schwefel's | Multimodal | 1 | 1 | $8.93 \times 10^{-4}$ | 206 | 1 | 0.001 | $6.68 \times 10^{-4}$ | 1422 |

of 0.01, 0.02, 0.03, 0.04 and 0.05 as mutation rates, then Group 2, consisting of 0.015, 0.04, 0.065, 0.09, 0.115 as mutation rates and Group 3 consisting of 0.025, 0.05, 0.075, 0.1, 0.125 as mutation rates. These mutation rates are applied to all the antibodies except for the first best four. In Genetic Algorithm, lower mutations rates are applied which are 0.005, 0.001 and 0.01 for each of the sets.

For each of the benchmark functions first the best clone rate/mutation rate combination are found for Clonal Selection and Genetic Algorithm and then these two rates are compared with proximity to the desired value (i.e. answer) as well as number of iterations to reach to the optimal solution. The comparison of algorithms with the best parameters is shown in Table 2. In all the functions both of the algorithms are applied and solutions are all found successfully. The iterations are all averaged over ten independent runs for each function and algorithm. If we consider function-by-function proximity to the actual answer is about the same so a correct determination of performance by analyzing the iterations is evaluated.

In sphere model Clonal Selection Algorithm reaches to a solution faster than the Genetic Algorithm. In Rastrigin's function the performance of Genetic Algorithm seems much better than the Clonal Selection Algorithm.

In Ackley's, Sum of Different Powers, and Non-generalized Schwefel's functions the Clonal Selection Algorithm works faster however for Modified Sinusoidal function Genetic Algorithm seems to work relatively faster.

## 5. CONCLUSIONS

In this work, Clonal Selection Algorithm and Genetic Algorithm were tested via different benchmark functions. Out of six benchmark functions; two benchmark functions were selected as a unimodal functions called smooth shaped functions, two of them were multimodal functions which has many local optimums, and two of them were highly multimodal functions which has much more local optimums than multimodal ones. Conclusions can be derived as looking at different types of functions.

Sphere and Sum of Different Powers are unimodal and convex functions. Ackley's and Non-generalized Schwefel's are multimodal functions, which has many local minimums. However, Rastrigin's and Modified Sinusoidal functions are highly multimodal functions which have much more local minimums than the multimodal functions. It is challenging to find the global optimum among the local optimums.

Clonal Selection Algorithm seems a better choice for a smooth shape, unimodal and multimodal functions like Sphere, Ackley's, Sum of Different Powers and Non-generalized Schwefel's functions. However, Genetic Algorithm has worked with better performance on highly multimodal test functions like Rastrigin's function and Modified Sinusoidal functions.

## ACKNOWLEDGEMENTS

The Authors would like to thank Kamil Dimililer and Ali Haydar from Girne American University for their suggestions and help with this work.

## Authors

**Ezgi Deniz Ülker** received the B.Sc. and M.Sc. degree in computer engineering from Girne American University, Girne, Cyprus, in 2008 and 2010 respectively, currently she is working towards her PhD degree in computer engineering. She worked as a teaching assistant in Girne American University between 2008-2010. She is working as a lecturer in Girne American University since September 2010. Her research interests include optimization techniques and studying different algorithms and their applications to solve different optimization and design problems.

**Sadık Ülker** received the B.Sc., M.E. and Ph.D. degree in electrical engineering from University of Virginia, Charlottesville, VA, USA, in 1996, 1999, and 2002 respectively.

He worked in the UVA Microwaves and Semiconductor Devices Laboratory between 1996-2001 as a research assistant. Later he worked in UVA Microwaves Lab as a Research Associate for one year. He joined Girne American University, Girne, Cyprus in September 2002. He was a member of electrical and electronics engineering department as an Assistant Professor until 2009. Since 2009, he has been promoted to Associate Professor status and he has also been the vice-rector of the Girne American University responsible from academic affairs. His research interests include Microwave Active Circuits, Microwave Measurement Techniques, Numerical Methods, and Metaheuristic Algorithms and their applications to design problems. He has authored and coauthored technical publications in the areas of submillimeter wavelength measurements and particle swarm optimization technique.

He is a member of IEEE and Eta Kappa Nu. He is also the recipient of Louis T. Rader Chairpersons award in 2001.